%% file: main.tex
\documentclass[11pt]{article}

\usepackage[margin=1in]{geometry}
\usepackage{newtxtext}
\usepackage{newtxmath}
\usepackage{microtype}
\usepackage{booktabs}
\usepackage{multirow}
\usepackage{graphicx}
\usepackage{tabularx}
\usepackage{makecell}
\usepackage{amsmath}
\usepackage{algorithm}
\usepackage{algorithmic}
\usepackage{xcolor}
\usepackage[hidelinks]{hyperref}
\usepackage[numbers,sort&compress]{natbib}

\title{Data-Efficient On-Policy Distillation for Automatic Speech Recognition}

\author{
Yu Lin \quad Yiming Wang \quad Runyuan Cai \quad Xiaodong Zeng\\
AutoArk-AI\\
\texttt{\{yu.lin,yiming.wang,runyuan.cai,xiaodong.zeng\}@autoark.ai}
}

\date{}

\begin{document}
\maketitle

\input{sections/abstract}
\input{sections/introduction}
\input{sections/related_work}
\input{sections/method}
\input{sections/experiments}
\input{sections/results}
\input{sections/discussion}
\input{sections/conclusion}

\bibliographystyle{unsrtnat}
\bibliography{references}

\end{document}

%% file: sections/abstract.tex
\begin{abstract}
Building competitive automatic speech recognition (ASR) models usually requires large-scale audio supervision, which makes reproduction and specialization expensive. We study Ark-ASR, a 0.6B-parameter audio-conditioned language model trained with 100k hours of speech, and examine whether a strong Qwen-ASR teacher can transfer additional recognition capability through on-policy distillation. Across Mandarin and English ASR benchmarks, the proposed training recipe consistently improves over supervised fine-tuning alone and outperforms the same-scale Qwen3-ASR-0.6B baseline on four of five evaluation sets. This is achieved with only 100k hours of speech, compared with the 20M hours of supervised audio reported for the Qwen3-Omni AuT encoder. The larger Qwen3-ASR-1.7B remains stronger, but the results show that teacher-guided on-policy training can substantially close the gap for compact ASR models under a much smaller audio budget. A support-overlap diagnostic further suggests that the teacher-data stage improves local student-teacher compatibility, matching recent analyses of when on-policy distillation is effective.
\end{abstract}

%% file: sections/introduction.tex
\section{Introduction}

Modern ASR has benefited from larger neural architectures, broader supervision, and increasingly capable pretrained audio models. End-to-end systems removed much of the hand-engineered pipeline used in earlier speech recognition systems \citep{hannun2014deepspeech}, while architectures such as Conformer improved the use of local and global acoustic context \citep{gulati2020conformer}. More recent systems scale supervision directly: Whisper demonstrates that training on 680k hours of multilingual weak supervision can produce robust zero-shot speech recognition \citep{radford2022whisper}, and Qwen3-Omni reports strong audio performance as part of a unified multimodal model family \citep{xu2025qwen3omni}. These trends improve accuracy, but they also make it difficult to train competitive ASR models when only a smaller curated audio budget is available.

Knowledge distillation offers a practical route for transferring behavior from a larger or stronger model into a smaller student \citep{hinton2015distilling}. In ASR, distillation is commonly implemented through pseudo-labels, sequence-level supervision, or teacher-generated training data, as in Distil-Whisper \citep{gandhi2023distilwhisper}. These approaches are effective when the teacher transcript is reliable, but they are mostly off-policy: the student is trained on static transcripts rather than on the errors and prefixes it currently visits during generation.

On-policy distillation (OPD) addresses this mismatch by training a student on its own generated trajectories with dense teacher feedback. Recent work on language-model OPD shows that OPD is not simply a free source of token-level supervision. Its success depends on local compatibility between student and teacher behavior, including alignment over high-probability token sets at student-visited states \citep{li2026rethinkingopd}. This observation is especially relevant to ASR, where a student may produce acoustically plausible but tokenization-sensitive transcripts and where teacher feedback must remain useful under the student's current transcript prefix.

This report studies Ark-ASR OPD, an ASR adaptation of OPD for a 0.6B-parameter student trained with 100k hours of audio. We use \emph{Ark-Base} for the 100k-hour supervised Ark-ASR checkpoint, and \emph{TD} for a teacher-data adaptation stage applied to Ark-Base before OPD. The method uses online student transcript generation, Qwen-ASR teacher scoring of the same audio and student text, and KL matching over a union support built from teacher top-$k$ tokens and student top-$k$ candidates. The main result is that Ark-Base+TD+OPD produces a 0.6B model that outperforms Qwen3-ASR-0.6B on four of five evaluation sets while using roughly 1/200 of the 20M-hour supervised audio scale reported for the Qwen3-Omni AuT encoder. The result does not close the gap to Qwen3-ASR-1.7B, but it shows that OPD can be a data-efficient route for improving compact ASR models when paired with a strong teacher and a compatible SFT initialization.

%% file: sections/related_work.tex
\section{Related Work}

\paragraph{Large-scale ASR training.}
End-to-end ASR systems have moved from specialized acoustic pipelines toward neural sequence models trained at increasing scale. Deep Speech showed the effectiveness of large end-to-end systems with substantial data and compute \citep{hannun2014deepspeech}. Conformer later combined convolutional and Transformer components for speech recognition \citep{gulati2020conformer}. Whisper further demonstrated that weakly supervised multilingual training at 680k hours can produce robust zero-shot performance across ASR benchmarks \citep{radford2022whisper}. Qwen3-Omni continues this scale-driven direction in a multimodal setting and reports strong audio-task performance across many audio and audio-visual benchmarks \citep{xu2025qwen3omni}. Ark-ASR OPD is complementary to these efforts: it studies how far a 0.6B ASR student can be pushed when the available training budget is 100k hours rather than the much larger scales used by frontier ASR systems.

\paragraph{Speech datasets and evaluation.}
The experiments use Mandarin and English ASR benchmarks that are widely used in ASR research. AISHELL-1 is an open-source Mandarin speech corpus with a Kaldi recipe and recognition baseline \citep{bu2017aishell}. WenetSpeech provides a much larger multi-domain Mandarin corpus and includes test sets such as Test\_Net and Test\_Meeting for matched internet and more difficult meeting conditions \citep{zhang2022wenetspeech}. LibriSpeech is a standard English ASR corpus derived from public-domain audiobooks and is commonly reported through test-clean and test-other splits \citep{panayotov2015librispeech}. We report CER for Mandarin sets and WER for LibriSpeech.

\paragraph{Knowledge distillation for ASR.}
Knowledge distillation transfers behavior from a stronger model or ensemble into a smaller model by matching softened predictions or generated labels \citep{hinton2015distilling}. In speech recognition, distillation is often expressed as pseudo-label training or teacher-guided sequence learning. Distil-Whisper uses large-scale pseudo-labeling to distill Whisper into a smaller model while preserving much of the teacher's robustness \citep{gandhi2023distilwhisper}. Other ASR distillation work studies transfer from autoregressive to non-autoregressive recognizers \citep{gong2022narasrkd}. Ark-ASR differs from static pseudo-label distillation by using the student's own online transcripts as the states at which teacher distributions are queried.

\paragraph{On-policy distillation.}
OPD has recently been studied as a post-training method for large language models. Rethinking OPD finds that successful OPD depends on compatible student-teacher behavior and progressive alignment over high-probability token sets at student-visited states \citep{li2026rethinkingopd}. Other contemporary OPD work studies variance control or local teachability failures in long-horizon settings \citep{oh2026vopd,liu2026prefixteach}. Ark-ASR OPD applies the same general principle to ASR: teacher supervision is computed on transcripts produced by the student, but the objective is adapted to audio-conditioned generation, tokenizer mapping, special-token masking, and top-$k$ support matching.

%% file: sections/method.tex
\section{Ark-ASR On-Policy Distillation}

\input{figures/ark_asr_architecture}
\input{figures/method_flow}

\subsection{Ark-ASR Architecture}

Ark-ASR is implemented as an audio-conditioned causal language model. The audio branch follows the GLM-ASR encoder design: a Whisper-style encoder converts speech features into acoustic hidden states, and a projector/MLP adapter maps those states into the language-model hidden space \citep{zai2025glmasr,huggingface2025glmasr}. In Ark-ASR, this branch adds layer normalization and temporal merging before the MLP adapter projects the acoustic states into the hidden dimension of the Qwen2 causal language model. At inference and training time, the adapted audio embeddings replace the embeddings at audio-token placeholder positions in the ASR prompt. The resulting mixed audio-text embedding sequence is processed by the causal decoder and LM head to generate transcript tokens autoregressively. Figure~\ref{fig:arkasr-architecture} summarizes this architecture.

\subsection{Training Flow}

Ark-ASR OPD trains an audio-capable causal language model student with a frozen Qwen-ASR teacher. For each audio batch, the student first generates a transcript without gradient tracking. The generated student token sequence is cleaned by removing ASR stop tokens and blocked non-ASR token ranges, then decoded into text. The teacher receives the same audio and the student transcript prefix in a teacher-forcing mode, producing token-level logits over the transcript positions. The student is then scored again with gradients under the same generated transcript so that its logits are aligned with the teacher's feedback at the states the student actually visited.

This differs from ordinary supervised fine-tuning. In SFT, the student is trained on reference or teacher-generated transcripts. In OPD, the teacher is not used as a static label source. It scores the student's current behavior, allowing dense feedback to target the distribution that the student induces during training.

\subsection{Union Top-$k$ KL Objective}

At each generated transcript position, the implementation constructs a local support set from two sources. The teacher contributes its top-$k$ tokens after mapping teacher-tokenizer ids to the student tokenizer. The student contributes its rollout top-$k$ candidates at the same position. The final support is the union of these token ids. Teacher log-probabilities are used for teacher-supported tokens, and teacher scores on student-only support tokens are gathered from the teacher-forced forward pass. The student logits are gathered on the same union support.

Let $U_t$ be the union support at transcript position $t$, $z^T_t$ the teacher scores on that support, and $z^S_t$ the student scores. With temperature $\tau$, Ark-ASR OPD minimizes
\begin{equation}
\mathcal{L}_{\mathrm{OPD}}
= \frac{1}{|\mathcal{T}|} \sum_{t \in \mathcal{T}}
\tau^2 \, \mathrm{KL}\left(
\mathrm{softmax}\left(z^T_t / \tau\right)
\middle\|
\mathrm{softmax}\left(z^S_t / \tau\right)
\right),
\end{equation}
where $\mathcal{T}$ contains generated transcript positions with at least two valid tokens in the union support. In the Qwen-ASR teacher-forcing path used in the experiments, the total training loss is this OPD loss.

\begin{algorithm}[t]
\caption{Ark-ASR OPD for one audio batch}
\label{alg:arkasr-opd}
\begin{algorithmic}[1]
\STATE Generate transcript tokens from the student without gradients.
\STATE Decode valid ASR tokens to text and infer the teacher language prompt.
\STATE Run the Qwen-ASR teacher in teacher-forcing mode on audio plus student text.
\STATE Collect teacher top-$k$ mapped student-token ids and teacher scores.
\STATE Re-score the same generated transcript with the student under gradients.
\STATE Build the union of teacher top-$k$ ids and student rollout top-$k$ ids.
\STATE Minimize temperature-scaled KL from teacher distribution to student distribution on the union support.
\end{algorithmic}
\end{algorithm}

\subsection{Practical ASR Details}

The ASR setting introduces several implementation details that are less visible in text-only OPD. Teacher and student tokenizers can differ, so token ids are mapped through shared token strings and invalid mappings are dropped. Special and control tokens are masked from teacher supports except for valid ASR stop behavior. Audio prompts must be kept aligned across batch padding, so the implementation tracks teacher alignment offsets and verifies retokenization mismatch rates. When a student rollout is empty, the implementation can fall back to teacher-forced reference text for scoring, but the reported runs show zero teacher-forced fallback in the analyzed logging window. The trainer uses FSDP2 over 24 workers and supports resumable checkpoints.

%% file: figures/ark_asr_architecture.tex
\begin{figure*}[t]
\centering
\includegraphics[width=\textwidth]{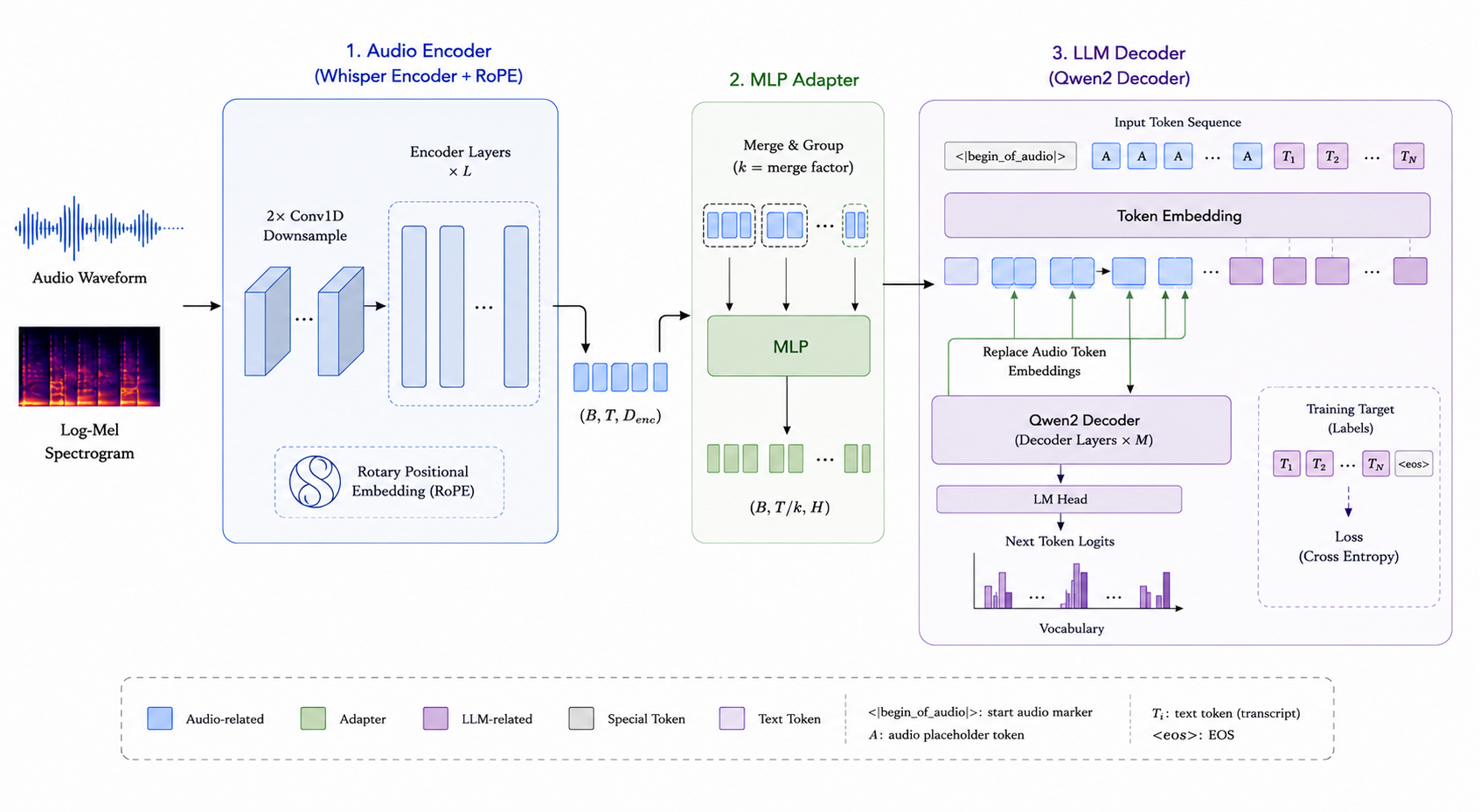}
\caption{Ark-ASR model architecture. The audio branch follows the GLM-ASR encoder design: a Whisper-style audio encoder produces frame-level acoustic states, which are normalized, temporally merged, and projected by an MLP adapter into the hidden dimension of the Qwen2 causal language model. These audio embeddings replace the placeholder audio-token embeddings in the ASR prompt, after which the causal decoder generates transcript tokens autoregressively.}
\label{fig:arkasr-architecture}
\end{figure*}

%% file: figures/method_flow.tex
\begin{figure*}[t]
\centering
\includegraphics[width=\textwidth]{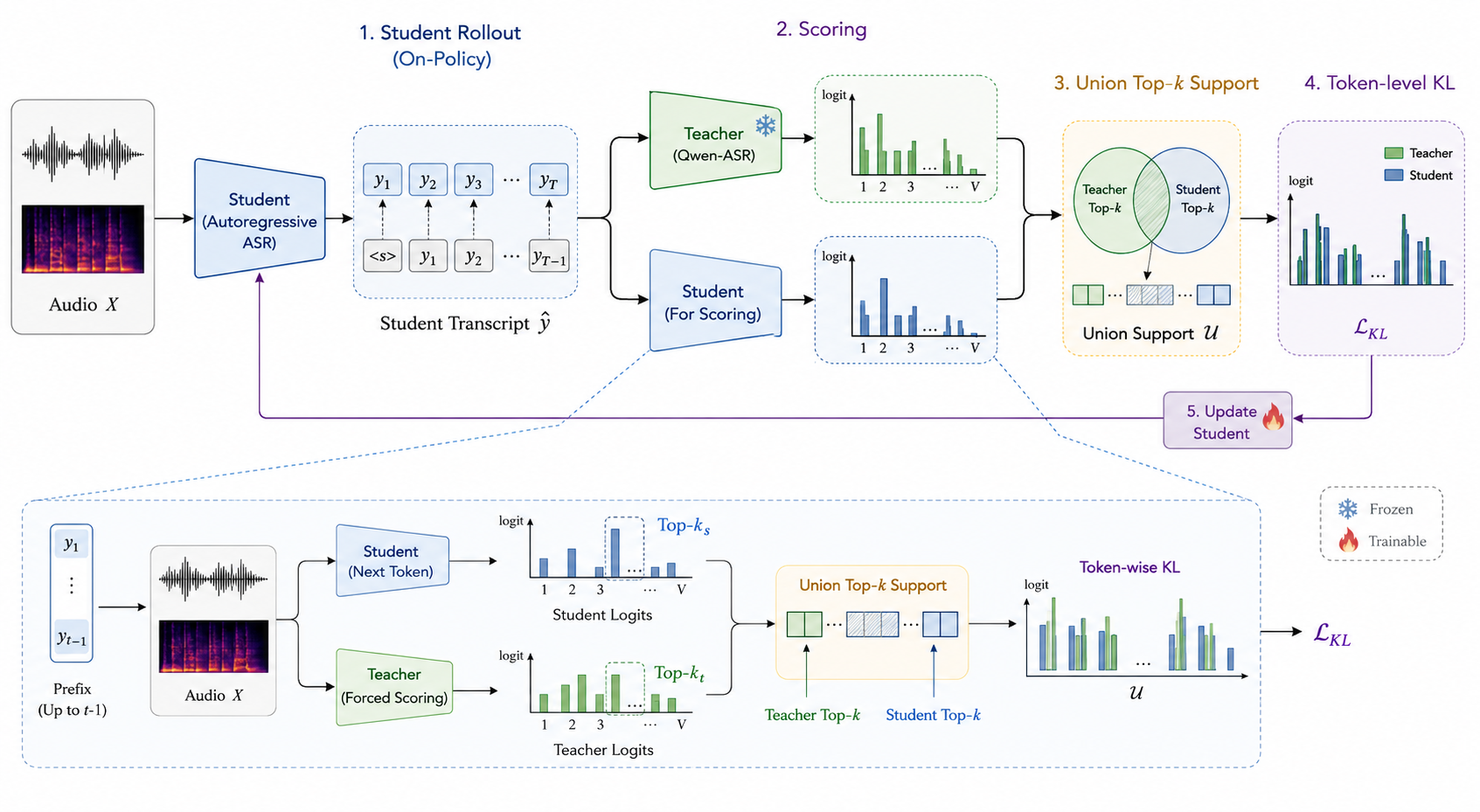}
\caption{Ark-ASR OPD training flow. The student generates transcripts on its own audio-conditioned states; the Qwen-ASR teacher scores the same audio and student text; the loss matches teacher and student distributions over the union of their local top-$k$ supports.}
\label{fig:method-flow}
\end{figure*}

%% file: sections/experiments.tex
\section{Experimental Setup}
\label{sec:experimental-setup}

\subsection{Models and Training Variants}

All Ark-ASR variants use a 0.6B-parameter student. The teacher for OPD is Qwen-ASR, loaded as a frozen scoring model. The training data scale for Ark-ASR is 100k hours. We compare three Ark-ASR recipes. \textbf{Ark-Base} denotes the 0.6B checkpoint obtained by supervised fine-tuning on the 100k-hour ASR data. \textbf{Ark-Base+OPD} starts from Ark-Base and performs Qwen-ASR OPD on the same 100k-hour dataset. \textbf{Ark-Base+TD+OPD} first applies a teacher-data (TD) adaptation stage to Ark-Base using 2,000 hours of teacher-generated ASR data, and then applies OPD. This third recipe is the strongest Ark-ASR setting in this report.

The baseline models are Qwen3-ASR-0.6B and Qwen3-ASR-1.7B. We use the Qwen3-ASR baselines as model-scale anchors rather than as a strict training-cost comparison, because the public technical report does not disclose all wall-clock training details. The relevant public comparison is scale and reported audio data: Qwen3-Omni reports an AuT encoder of about 0.6B parameters trained on 20M hours of supervised audio \citep{xu2025qwen3omni}, whereas Ark-ASR in this report uses 100k hours for the reported OPD experiments.

\subsection{Evaluation}

We evaluate on five ASR test sets. AISHELL-1, WenetSpeech Test\_Meeting, and WenetSpeech Test\_Net are reported with character error rate (CER). LibriSpeech test-clean and test-other are reported with word error rate (WER). Lower values are better for both metrics. The evaluation script normalizes text, removes punctuation for scoring, and computes edit-distance based CER/WER after inference.

\subsection{Training Diagnostics}

To inspect the mechanism behind the strongest recipe, we report Valid Union Support Size (VUSS), a support-overlap diagnostic for OPD. VUSS is the number of valid tokens retained after merging teacher and student top-$k$ candidate sets and filtering invalid token mappings at student-visited transcript states. We compare OPD convergence behavior before and after the TD stage. This diagnostic is not used as a benchmark metric; it is used to understand whether the student and teacher have more compatible local token supports after TD adaptation.

%% file: sections/results.tex
\section{Results and Analysis}

\subsection{Main ASR Results}

\input{tables/main_results}

Table~\ref{tab:main-results} shows that Ark-Base alone is not sufficient to match the Qwen3-ASR-0.6B baseline, but it provides a useful initialization for OPD. Ark-Base+OPD improves every benchmark: AISHELL-1 improves from 3.48\% to 3.00\% CER, WenetSpeech meeting from 10.22\% to 7.18\% CER, WenetSpeech net from 7.74\% to 6.13\% CER, LibriSpeech clean from 3.75\% to 2.88\% WER, and LibriSpeech other from 7.17\% to 5.50\% WER.

Ark-Base+TD+OPD is the strongest Ark-ASR recipe. It reaches 1.95\% CER on AISHELL-1, 5.92\% and 5.39\% CER on WenetSpeech meeting/net, and 2.45\% and 4.56\% WER on LibriSpeech clean/other. At the same 0.6B model scale, this final Ark-ASR model is stronger overall than Qwen3-ASR-0.6B, improving AISHELL-1, WenetSpeech net, LibriSpeech clean, and LibriSpeech other. Qwen3-ASR-1.7B remains the best model in the table, indicating that the current 0.6B recipe does not remove the advantage of the larger model.

\subsection{Top-$k$ Support Compatibility}

\input{tables/topk_diagnostics}

Table~\ref{tab:topk-diagnostics} summarizes VUSS during OPD convergence. Before the TD stage, the mean VUSS is 53.06. After the TD stage, the corresponding value is 51.61. Because VUSS measures the size of the valid union support formed from teacher and student top-$k$ candidates, a smaller value under the same top-$k$ setting indicates that the two supports have become more overlapping or locally compatible.

This pattern matches the performance trend. The model after TD adaptation obtains better final CER/WER after OPD, and its lower VUSS suggests a more compatible local support between student and teacher. The observation is also consistent with recent OPD analysis, which argues that effective OPD depends on compatible student-teacher behavior and alignment on high-probability token sets at student-visited states \citep{li2026rethinkingopd}. The diagnostic should not be read as a complete causal proof, since it compares two training recipes with different initialization stages, but it supports the interpretation that teacher-data adaptation makes subsequent OPD supervision more locally usable.

%% file: tables/main_results.tex
\begin{table*}[t]
\centering
\small
\setlength{\tabcolsep}{3.5pt}
\newcommand{\best}[1]{\ensuremath{\mathbf{#1}}}
\begin{tabularx}{\textwidth}{Xccccc}
\toprule
Model & AISHELL-1 & \makecell{Wenet\\Meeting} & \makecell{Wenet\\Net} & \makecell{Libri\\Clean} & \makecell{Libri\\Other} \\
\midrule
\multicolumn{6}{l}{\textit{0.6B models}} \\
Ark-Base & 3.48 & 10.22 & 7.74 & 3.75 & 7.17 \\
Ark-Base+OPD & 3.00 & 7.18 & 6.13 & 2.88 & 5.50 \\
Ark-Base+TD+OPD & \best{1.95} & 5.92 & \best{5.39} & \best{2.45} & \best{4.56} \\
Qwen3-ASR-0.6B & 2.07 & \best{5.57} & 5.45 & 2.81 & 5.05 \\
\midrule
\multicolumn{6}{l}{\textit{Larger reference model}} \\
Qwen3-ASR-1.7B & 1.50 & 4.69 & 4.55 & 2.20 & 4.05 \\
\bottomrule
\end{tabularx}
\caption{ASR evaluation results. Mandarin datasets are reported with CER and LibriSpeech datasets with WER. Lower is better. Bold numbers mark the best result within the 0.6B group. Ark-Base is the 100k-hour supervised checkpoint; TD denotes teacher-data adaptation; OPD denotes on-policy distillation.}
\label{tab:main-results}
\end{table*}

%% file: tables/topk_diagnostics.tex
\begin{table}[t]
\centering
\small
\begin{tabular}{lccc}
\toprule
Condition & Mean & Min & Max \\
\midrule
Before TD stage & 53.06 & 52.38 & 54.20 \\
After TD stage & 51.61 & 50.88 & 52.41 \\
\bottomrule
\end{tabular}
\caption{Valid Union Support Size (VUSS) during OPD convergence. VUSS measures the number of valid tokens retained after merging teacher and student top-$k$ candidate sets and filtering invalid token mappings. Under the same top-$k$ setting, a smaller VUSS is consistent with greater overlap between teacher and student local supports.}
\label{tab:topk-diagnostics}
\end{table}

%% file: sections/discussion.tex
\section{Discussion and Limitations}

The results suggest that OPD is most useful when the student is already close enough to the teacher for dense token-level feedback to be locally meaningful. Ark-Base creates a workable ASR student from 100k hours of SFT data, and OPD improves it across all five test sets. The TD stage further improves the initialization, after which OPD gives the best 0.6B model in the comparison. This sequence follows the mechanism suggested by OPD studies in language modeling: student-teacher compatibility matters because the teacher must provide discriminative feedback on states the student actually visits \citep{li2026rethinkingopd}.

The comparison should be interpreted with care. Qwen3-ASR-1.7B is both larger and part of a different training pipeline, so it is not a controlled ablation of model size or data alone. Qwen3-ASR-0.6B is a useful same-scale baseline, but its full training recipe is not reproduced here. The Ark-ASR result therefore supports a practical claim about data-efficient improvement under the available 100k-hour setup, not a claim that OPD universally replaces large-scale ASR pretraining. The 20M-hour figure used for scale context is taken from the Qwen3-Omni report's AuT encoder description, not from an independently reproduced Qwen3-ASR training run.

The current experiments also leave several open questions. The paper does not report repeated seeds, per-domain data composition, or compute-normalized training cost. The VUSS analysis is a diagnostic rather than a controlled intervention over support overlap. Future work should vary the SFT initialization, teacher strength, OPD top-$k$, and rollout length while holding data constant. A more complete study should also measure long-form robustness, hallucination behavior, and streaming latency, which are important for deployed ASR but outside the scope of this short report.

%% file: sections/conclusion.tex
\section{Conclusion}

This report presents Ark-ASR OPD, an adaptation of on-policy distillation to automatic speech recognition. The method trains a 0.6B ASR student on its own audio-conditioned transcript rollouts using dense Qwen-ASR teacher feedback over a union top-$k$ support. With only 100k hours of training audio, Ark-Base followed by OPD improves all evaluated benchmarks, and Ark-Base+TD+OPD surpasses Qwen3-ASR-0.6B overall while remaining below Qwen3-ASR-1.7B. The training diagnostics indicate that the TD stage makes the student and teacher more locally compatible under the OPD support metric, matching recent findings on when OPD is effective. These results point to SFT plus OPD as a practical, data-efficient path for improving compact ASR models.

%% file: references.bib
@article{hinton2015distilling,
  title={Distilling the Knowledge in a Neural Network},
  author={Hinton, Geoffrey and Vinyals, Oriol and Dean, Jeff},
  journal={arXiv preprint arXiv:1503.02531},
  year={2015},
  doi={10.48550/arXiv.1503.02531},
  url={https://arxiv.org/abs/1503.02531}
}

@article{hannun2014deepspeech,
  title={Deep Speech: Scaling up end-to-end speech recognition},
  author={Hannun, Awni and Case, Carl and Casper, Jared and Catanzaro, Bryan and Diamos, Greg and Elsen, Erich and Prenger, Ryan and Satheesh, Sanjeev and Sengupta, Shubho and Coates, Adam and Ng, Andrew Y.},
  journal={arXiv preprint arXiv:1412.5567},
  year={2014},
  doi={10.48550/arXiv.1412.5567},
  url={https://arxiv.org/abs/1412.5567}
}

@article{gulati2020conformer,
  title={Conformer: Convolution-augmented Transformer for Speech Recognition},
  author={Gulati, Anmol and Qin, James and Chiu, Chung-Cheng and Parmar, Niki and Zhang, Yu and Yu, Jiahui and Han, Wei and Wang, Shibo and Zhang, Zhengdong and Wu, Yonghui and Pang, Ruoming},
  journal={arXiv preprint arXiv:2005.08100},
  year={2020},
  doi={10.48550/arXiv.2005.08100},
  url={https://arxiv.org/abs/2005.08100}
}

@inproceedings{panayotov2015librispeech,
  title={LibriSpeech: An ASR corpus based on public domain audio books},
  author={Panayotov, Vassil and Chen, Guoguo and Povey, Daniel and Khudanpur, Sanjeev},
  booktitle={2015 IEEE International Conference on Acoustics, Speech and Signal Processing (ICASSP)},
  pages={5206--5210},
  year={2015},
  doi={10.1109/ICASSP.2015.7178964},
  url={https://doi.org/10.1109/ICASSP.2015.7178964}
}

@article{bu2017aishell,
  title={AISHELL-1: An Open-Source Mandarin Speech Corpus and A Speech Recognition Baseline},
  author={Bu, Hui and Du, Jiayu and Na, Xingyu and Wu, Bengu and Zheng, Hao},
  journal={arXiv preprint arXiv:1709.05522},
  year={2017},
  doi={10.48550/arXiv.1709.05522},
  url={https://arxiv.org/abs/1709.05522}
}

@article{zhang2022wenetspeech,
  title={WenetSpeech: A 10000+ Hours Multi-domain Mandarin Corpus for Speech Recognition},
  author={Zhang, Binbin and Lv, Hang and Guo, Pengcheng and Shao, Qijie and Yang, Chao and Xie, Lei and Xu, Xin and Bu, Hui and Chen, Xiaoyu and Zeng, Chenchen and Wu, Di and Peng, Zhendong},
  journal={arXiv preprint arXiv:2110.03370},
  year={2022},
  doi={10.48550/arXiv.2110.03370},
  url={https://arxiv.org/abs/2110.03370}
}

@article{radford2022whisper,
  title={Robust Speech Recognition via Large-Scale Weak Supervision},
  author={Radford, Alec and Kim, Jong Wook and Xu, Tao and Brockman, Greg and McLeavey, Christine and Sutskever, Ilya},
  journal={arXiv preprint arXiv:2212.04356},
  year={2022},
  doi={10.48550/arXiv.2212.04356},
  url={https://arxiv.org/abs/2212.04356}
}

@article{gandhi2023distilwhisper,
  title={Distil-Whisper: Robust Knowledge Distillation via Large-Scale Pseudo Labelling},
  author={Gandhi, Sanchit and von Platen, Patrick and Rush, Alexander M.},
  journal={arXiv preprint arXiv:2311.00430},
  year={2023},
  doi={10.48550/arXiv.2311.00430},
  url={https://arxiv.org/abs/2311.00430}
}

@article{gong2022narasrkd,
  title={Knowledge Transfer and Distillation from Autoregressive to Non-Autoregressive Speech Recognition},
  author={Gong, Xun and Zhou, Zhikai and Qian, Yanmin},
  journal={arXiv preprint arXiv:2207.10600},
  year={2022},
  doi={10.48550/arXiv.2207.10600},
  url={https://arxiv.org/abs/2207.10600}
}

@article{xu2025qwen3omni,
  title={Qwen3-Omni Technical Report},
  author={Xu, Jin and Guo, Zhifang and Hu, Hangrui and Chu, Yunfei and Wang, Xiong and He, Jinzheng and Wang, Yuxuan and Shi, Xian and He, Ting and Zhu, Xinfa and others},
  journal={arXiv preprint arXiv:2509.17765},
  year={2025},
  doi={10.48550/arXiv.2509.17765},
  url={https://arxiv.org/abs/2509.17765}
}

@article{li2026rethinkingopd,
  title={Rethinking On-Policy Distillation of Large Language Models: Phenomenology, Mechanism, and Recipe},
  author={Li, Yaxuan and Zuo, Yuxin and He, Bingxiang and Zhang, Jinqian and Xiao, Chaojun and Qian, Cheng and Yu, Tianyu and Gao, Huan-ang and Yang, Wenkai and Liu, Zhiyuan and Ding, Ning},
  journal={arXiv preprint arXiv:2604.13016},
  year={2026},
  doi={10.48550/arXiv.2604.13016},
  url={https://arxiv.org/abs/2604.13016}
}

@article{oh2026vopd,
  title={KL for a KL: On-Policy Distillation with Control Variate Baseline},
  author={Oh, Minjae and Song, Sangjun and Choi, Gyubin and Choi, Yunho and Jo, Yohan},
  journal={arXiv preprint arXiv:2605.07865},
  year={2026},
  doi={10.48550/arXiv.2605.07865},
  url={https://arxiv.org/abs/2605.07865}
}

@article{liu2026prefixteach,
  title={Prefix Teach, Suffix Fade: Local Teachability Collapse in Strong-to-Weak On-Policy Distillation},
  author={Liu, Kaiyuan and Zhuang, Ziyuan and Bai, Yang and Wang, Bing and Weng, Rongxiang and Ye, Jieping},
  journal={arXiv preprint arXiv:2605.13643},
  year={2026},
  doi={10.48550/arXiv.2605.13643},
  url={https://arxiv.org/abs/2605.13643}
}

@misc{zai2025glmasr,
  title={{GLM-ASR}: A Robust, Open-Source Speech Recognition Model},
  author={{Z.ai}},
  year={2025},
  howpublished={GitHub repository},
  url={https://github.com/zai-org/GLM-ASR}
}

@misc{huggingface2025glmasr,
  title={{GlmAsr} Model Documentation},
  author={{Hugging Face}},
  year={2025},
  howpublished={Transformers documentation},
  url={https://huggingface.co/docs/transformers/model_doc/glmasr}
}
